\documentclass[conference]{IEEEtran}
\IEEEoverridecommandlockouts
\usepackage{cite}
\usepackage{amsmath,amssymb,amsfonts}
\usepackage{algorithmic}
\usepackage{graphicx}
\usepackage{textcomp}
\def\BibTeX{{\rm B\kern-.05em{\sc i\kern-.025em b}\kern-.08em
    T\kern-.1667em\lower.7ex\hbox{E}\kern-.125emX}}

\usepackage[skip=3pt]{caption}
\usepackage{tikz}
\usepackage{textcomp}
\usepackage{hyperref}
\usepackage{lipsum}

\newcommand\copyrighttext{%
	\footnotesize \textcopyright 2017 IEEE. Personal use of this material is permitted. Permission from IEEE must be obtained for all other uses, in any current or future media, including reprinting/republishing this material for advertising or promotional purposes, creating new collective works, for resale or redistribution to servers or lists, or reuse of any copyrighted component of this work in other works.
}
\newcommand\copyrightnotice{%
	\begin{tikzpicture}[remember picture,overlay]
	\node[anchor=south,yshift=10pt] at (current page.south) {\fbox{\parbox{\dimexpr\textwidth-\fboxsep-\fboxrule\relax}{\copyrighttext}}};
	\end{tikzpicture}%
}
\begin{document}

\title{Learning and  Real-time Classification of Hand-written Digits   With  Spiking Neural Networks 
\thanks{This research was supported in part by grants from  National Science Foundation Award 1710009, CISCO, and the McNair Fellowship Program.}
}

\author{\IEEEauthorblockN{Shruti R. Kulkarni, John M. Alexiades, Bipin Rajendran}
\IEEEauthorblockA{Department of Electrical and Computer Engineering, \\New Jersey Institute of Technology, Newark, NJ, 07102, USA\\
Email: \{srk68, jma59, bipin\}@njit.edu}
}

\maketitle
\copyrightnotice
\begin{abstract}

We describe a novel spiking neural network (SNN) for automated, real-time handwritten digit classification and its implementation on a GP-GPU platform.  Information processing within the  network, from feature extraction to classification is implemented by mimicking the basic aspects of neuronal spike initiation and propagation in the brain. The feature extraction layer of the SNN uses fixed synaptic weight maps to extract the key features of the image and the classifier layer uses the recently developed NormAD
approximate gradient descent based supervised learning algorithm for spiking neural networks to adjust the synaptic weights. On the standard MNIST database  images of handwritten digits, our network achieves an accuracy of $99.80\%$ on the training set and $98.06\%$ on the test set, with nearly $7\times$ fewer parameters compared to the state-of-the-art spiking networks. We further use this  network in a GPU based user-interface system demonstrating real-time SNN simulation to infer digits written by different users. On a test set of $500$ such images, this real-time platform  achieves an accuracy exceeding $97\%$ while making a prediction within an SNN emulation time of less than $100\,$ms.
\end{abstract}


\section{Introduction}
The human brain is a computational marvel compared to man-made  systems, both in its ability to learn to execute highly complex cognitive tasks, as well as in its  energy efficiency. The computational efficiency of the brain stems from its use of   sparsely issued binary  signals or spikes to encode and process information. Inspired by this, spiking neural networks (SNNs) have been proposed as a computational framework  for  learning and inference \cite{snn1}. General purpose graphical processing units (GP-GPUs) have become an ideal platform for accelerated implementation of  large scale machine learning algorithms~\cite{COTS}. 
There have been multiple GPU based implementations for simulating large SNNs \cite{nemo, gpu_snn, genn, gpu_sim_snn,naveros2017event,krichmar2015large}, with most of these targeting the forward communication of spikes through  large  networks of spiking neurons and/or local weight update based on spike timing difference. In contrast, we demonstrate a highly optimized real time implementation scheme  for spike based supervised learning on GPU platforms and use the  framework for real time inference on digits captured from different users through a touch-screen interface.

Previous efforts to develop deep convolutional spiking networks started by using  second generation artificial neural networks (ANNs) with   back-propagation of errors to train the network and thereafter converting it into  spiking versions \cite{Diehl,Energyeff,rueckauer2016theory,hunsberger2016training}. There have been several supervised learning algorithms proposed to train the SNNs, by explicitly using the time of spikes of neurons to encode information, and to derive the appropriate weight update rules to minimize the distance between desired spike times and observed spike times in a network \cite{NormAD,resume,span,snn_bp,cone}.
We use the  Normalized Approximate Descent (NormAD) algorithm to design a system to identify handwritten digits. The NormAD algorithm has shown superior convergence speed compared to other methods such as the Remote Supervised Method (ReSuMe) \cite{NormAD}. 

Our SNN is trained on the  MNIST database consisting of $60,000$ training images  and $10,000$ test images 
\cite{mnist}. 
The highest accuracy SNN for the MNIST  was reported in \cite{snn_bp}, where  a two-stage convolution neural network achieved an accuracy of $99.31\%$ on the  test set. Our network, in contrast, has just three layers, with about  $82,000$ learning synapses ($7\times$ fewer parameters compared to \cite{snn_bp}) and achieves an accuracy of $98.06\%$ on the MNIST test dataset. 

The paper is organized as follows. The  computational units of the SNN and the network architecture are described in section \ref{snn_arch}. Section \ref{cuda} details how the network simulation is divided among different CUDA kernels. 
The user-interface system and the image pre-processing steps are explained in  Section \ref{preprocess}. We present the results of our network simulation and speed related optimizations in Section \ref{res}. Section \ref{concl} concludes our GPU based system implementation  study.

\section{Spiking Neural Network}
\label{snn_arch}
The basic units of an SNN are  spiking neurons and synapses interconnecting them. 
For computational tractability,  we use the leaky integrate and fire (LIF) model of  neurons, where the evolution of the   membrane potential, $V_m(t)$  is described by: %the   differential equation,
\begin{equation}
C\frac{dV_m(t)}{dt}=-g_L(V_m(t)-E_L)+I(t)
\label{eq:lif}
\end{equation}
Here $I(t)$ is the total input current, $E_L$ is the resting potential, and  $C$ ($300\,$pF) and $g_L$ ($30\,$nS) model the  membrane capacitance and leak conductance, respectively \cite{NormAD}. 
Once the membrane potential crosses a threshold ($V_m(t)  \ge V_T$), it is reset to its resting value $E_L$ and remains at that value till the neuron comes out of its refractory period ($t_{ref}=3\,$ms).  
The synapse, with weight $w_{k,l}$  connecting input neuron $k$ to output neuron $l$, transforms the incoming spikes (arriving at times $t_{k}^1,t_{k}^2,\ldots$) into a post-synaptic current ($I_{k,l}$), based on the following transformation,
\begin{align}
&c_k(t)=\sum_{{i}} \delta(t-t_{k}^i)*\left(e^{ -t/\tau_1 }-e^{ -t/\tau_2 }\right)
\label{eq:ci_kernel}\\
&I_{k,l} (t) = w_{k,l} \times c_k(t)
\label{eq:syn_curr}
\end{align}
Here, the summed $\delta$ function represents the incoming spike train and the double decaying exponentials with $\tau_1=5\,$ms  and $\tau_2=1.25\,$ms represent the synaptic kernel. These  values  closely match the biological time constants \cite{dayan2003theoretical}.

\subsection{Network architecture \& spike encoding}
\begin{figure}[!h]
	\centering
	\includegraphics[height=0.49\linewidth]{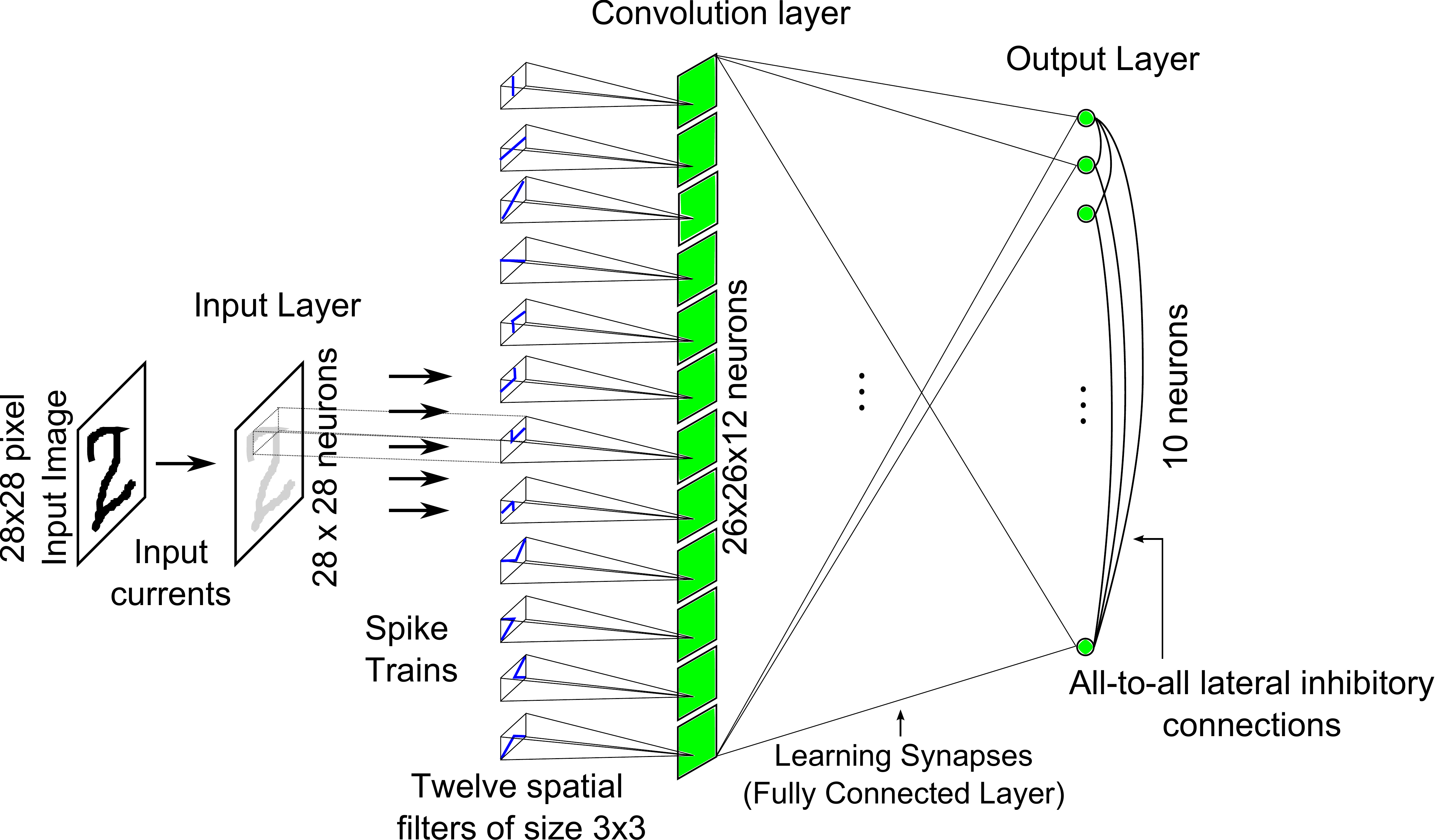}\\
	\vspace{.1in}
    \includegraphics[width=0.43\textwidth]{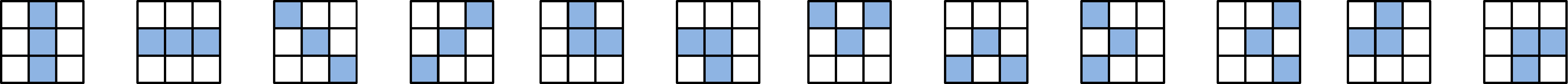}
	\caption{The $28\times 28$ pixel images from the MNIST database are  converted to spike trains, which are presented for a duration $T$,  weighted with twelve $3\times 3$ synaptic weight maps (below) resulting in twelve  $26\times 26$ current streams and then feed to  the   corresponding feature map neurons. There are $10$ output neurons corresponding to each digit. %The output layer fixed strength lateral connections among all the $10$ neurons, which provide inhibitory currents. 
The weights of the fully-connected feed-forward synapses to the output layer neurons ($8112\times 10$) are adjusted using the NormAD learning rule \cite{NormAD}. Additionally, the output layer neurons also have lateral inhibitory connections. } 
	\label{fig:mnist_arch}
\end{figure}

We use a three-layered network where hidden layer performs feature extraction and the output layer performs classification (see Fig.~\ref{fig:mnist_arch}). The network is designed to take input from  $28\times28$  pixel  MNIST digit image. 
We translate this pixel value into a set of spike streams, by passing the pixels as currents to a layer of $28\times28$ neurons (first layer). 
The current  $i(k)$  applied to  a neuron corresponding to pixel value $k$, in the range $[0,255]$ is obtained by the following linear relation:
\begin{equation}
i(k)=I_0+k\times I_p
\end{equation}
where $I_p=101.2\,$pA is a scaling factor,  and $I_0=2700\,$pA is the minimum current above which an LIF neuron can generate a spike (for the parameters   chosen in equation \ref{eq:lif}). %and $w_b$ is the weighting factor for each pixel.
These spike streams are then weighted with twelve $3\times 3$ synaptic weight maps (or filters), with \textit{a priori} chosen values  to generate equivalent  current streams using equations \ref{eq:ci_kernel} and  \ref{eq:syn_curr}. These $12$  spatial filter  maps  are chosen to detect various edges and corners in the image. 

The output layer consists of $10$ neurons, one for each of the ten digits. 
We train the network so that the correct neuron in the output layer generates a spike train with a frequency close to $285\,$Hz and the other output neurons issue no spikes during the presentation duration, $T$ (set to $100\,$ms in baseline experiments). $T$ is also a hyper-parameter of our network, and its effect on the network's classification ability will be discussed in section \ref{res}.  This layer also has lateral inhibitory connections that helps to prevent the non-label neurons from spiking for a given input. The output neuron with the highest number of spikes is declared the winner of the classification.

\subsection{Learning layer}
The $8112\times10$ synapses connecting the hidden layer neurons to the $10$ output layer neurons are modified during the course of training using the NormAD rule \cite{NormAD}. The strength of the weights are adjusted based on the error between the observed and desired spike streams ($e(t)=S^{d}(t)-S^{o}(t)$) and the term $\hat{d}(t)$, denoting the effect of incoming spike kernels on the neuron's membrane potential, according to the relation: 
%The  NormAD \cite{NormAD} supervised learning algorithm is used to calculate the  weight  update:
\begin{align}
\Delta \mathbf{w}& = r\int_{0}^{T} e(t)\frac{\mathbf{\hat{d}}(t)}{\| \mathbf{\hat{d}}(t)\|} dt,&\mathbf{\hat{d}}(t) &= \mathbf{c}(t) * \hat{h}(t)
\label{eq:normad_wt}
\end{align}
where $\hat{h}(t)=(1/C)\exp(-t/\tau_{L})$ represents the neuron's impulse response with $\tau_L=1\,$ms, and $r$ is the learning rate.

\section{CUDA implementation}
\label{cuda}
The SNN is implemented on a GPU platform using the CUDA-C programming framework. A GPU is divided into streaming multiprocessors (SM), each of which consists of stream processors (SP) that are optimized to execute math operations. The CUDA-C  programming framework   exploits the hardware parallelism of GPUs and launches jobs on the GPU in a grid of blocks each mapped to an SM. The blocks are further divided into multiple threads, each of which is scheduled to run on an SP, also called a CUDA core. %To offload the computation on the GPU, memory needs to be assigned on the GPU memory for   network variables. 
Since memory transfer between CPU and GPU local memory is one of the main bottlenecks,  all  network variables (i.e., neuron membrane potentials and synaptic currents) are declared in the global GPU memory in our implementation. 
The simulation equations (\ref{eq:lif}), (\ref{eq:ci_kernel}) and (\ref{eq:syn_curr}) are %solved 
evaluated numerically in an iterative manner at each time step.

\begin{figure}[!h]
	\centering
	\includegraphics[width=0.47\textwidth]{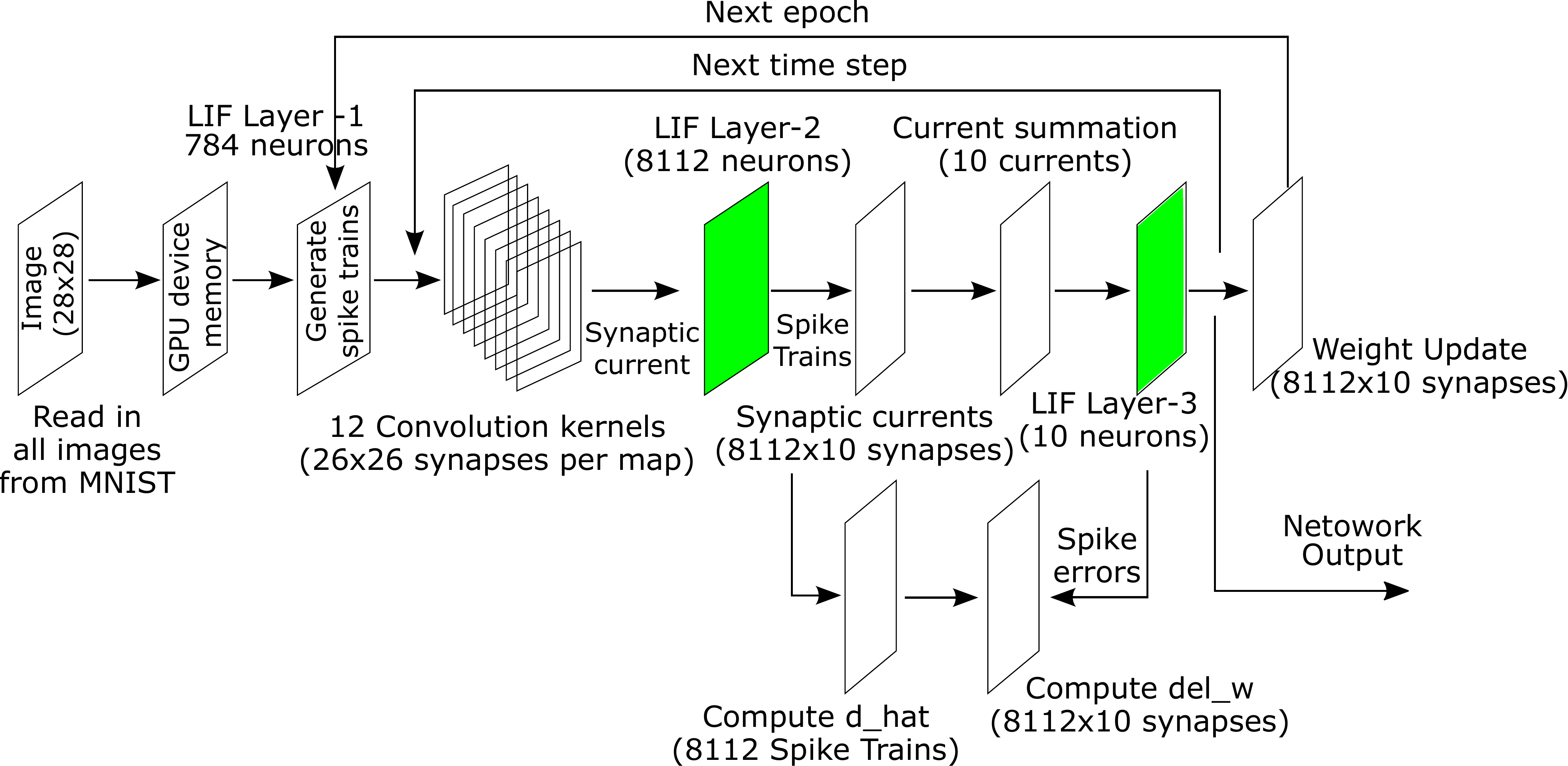}
	\caption{Diagram showing the different variables of the network being computed each time step and how the signals flow across different layers. The dimensions within the brackets are the sizes of those variables and their respective CUDA kernels. } 
	\label{fig:training_scheme}
\end{figure}

Fig.~\ref{fig:training_scheme} shows the forward pass and backward pass for weight update during the training phase. Image pixels read into the GPU memory are passed as currents to layer one neurons (grid size of $28\times28$) for the presentation duration, $T$. The filtering process involves 2D convolution of the incoming spike kernels and the weight matrix ($3\times3$). The computation is parallelized across $12$ CUDA kernels, each with a grid size of $26\times 26$ threads. Each thread computes the current to the hidden layer neurons, indexed as a 2D-array $i,j$, $\{0\leq i,j,\leq25\}$ at a time-step $n$, based on the following spatial convolution relation:
\begin{align}
I_{in}(i,j,n)=\sum_{a=0}^{2}\sum_{b=0}^{2}w_{conv}(a,b)\times c(i+a,j+b,n)
\label{eq:conv}
\end{align}
where $c$ represents the synaptic kernel (equation \ref{eq:ci_kernel}) calculated from the spike trains of the $28\times 28$ pixels and $w_{conv}(a,b)$ represents each of the weights from the $3\times3$ filter matrix. 

The membrane potential of an array of $k$ LIF neurons, for applied current $\mathbf{I}(n)$ (as described in equation \ref{eq:lif}) is evaluated using the second order Runge-Kutta method as: 
\begin{align}
&\mathbf{k_1}= [-g_L(\mathbf{V_m}(n)-E_L)+\mathbf{I}(n)]/C\\
&\mathbf{k_2}= [-g_L(\mathbf{V_m}(n) +\mathbf{k_1}\Delta t-E_L)+\mathbf{I}(n)]/C\\
&\mathbf{V_m}(n+1)  =  \mathbf{V_m}(n)  + [(\mathbf{k_1}+\mathbf{k_2})\Delta t/2]
\label{eq:RK_lif}
\end{align}
Each thread $k$ independently checks if the membrane potential has exceeded the threshold to artificially reset it. % to the resting potential.
\begin{align}
\text{If } V_{m}^k(n+1)&\geq V_T\Rightarrow V_{m}^k(n+1) =E_L
\end{align}
Refractory period is implemented by storing the latest spike issue time, $n_k^{last}$ of each neuron in a  vector $\mathbf{R}$;  the membrane potential of a neuron is updated only when the current time step $n>n_k^{last}+(t_{ref}/\Delta t)$.

The synaptic current   from  neuron $k$ in hidden layer to    neuron $l$ in output layer as given in equation \ref{eq:syn_curr} can be re-written to be evaluated in an iterative manner, thereby avoiding the evaluation of expensive exponential of the difference between the current time $n$ and previous spike times $n_k^i$. The synaptic current computation, at  time step $n$, for each of the $(k,l)$ synapse is spawned in CUDA across $8112\times 10$ kernels as: 
\begin{align}
&a_{k}(n) =  a_{k}(n-1)\times \exp(-\Delta t/\tau_1) + \delta(n-n_k^i)\\
&b_{k}(n) =  b_{k}(n-1)\times \exp(-\Delta t/\tau_2) + \delta(n-n_k^i)\\
&c_{k}(n) =  a_{k}(n)-b_{k}(n)\\
&I_{{k,l}}=  w_{k,l}\times c_{k}(n) 
\end{align}
where $a_k(n)$ and $b_k(n)$ represent the rising and falling regions of the double exponential synaptic kernel. The strength of the synapses between the hidden and output layers is initialized to zero during training.  
At every time step, the error function for each output neuron is calculated, based on the difference between the observed and desired spikes. Next, $\hat{d}_k$  (equation \ref{eq:normad_wt}) for the spikes originating from neuron $k$ is computed as:
\begin{equation}
\hat{d}_{k}(n)=\hat{d}_{k}(n-1)e^{-\Delta t/\tau_L}+(c_{k}(n)  \Delta t)/C
\end{equation}

Once $\hat{d}_{k}(n)$ is evaluated, we compute its norm across all $k$ neurons and determine the instantaneous $\Delta w_{k,l}(n)$ for all the $81,120$ synapses in parallel, if there is a spike error. At the end of  presentation, the accumulated $\Delta w_{k,l}$ is used to update the synaptic weights  in parallel. The evaluation of the total synaptic current and the norm is performed using parallel reduction in CUDA \cite{Nvidia_docs}. During the inference or testing phase, we  calculate the synaptic currents and membrane potentials of neurons in both layers to determine spike times, but do not evaluate the $\mathbf{\hat{d}}$ term and the weight update.  %\color{red}

%\begin{figure}[!htb]
%	\centering%
%	\includegraphics[width=0.4\linewidth]{Display.png}
%	\caption{Comparison of user input (left) to finalized 28x28 input image (right)
%	}    
%	\label{fig:setup}
%\end{figure}

%Our CUDA implementation on the K80 GPU achieved a speedup of close to $6.7\times$ for learning and  $5.2\times$ improvement for inference compared to a single core CPU based simulation on a Xeon-E5 processor. It can be seen that during the learning phase where the weight update CUDA kernels are called, the overall speedup (with respect to CPU) is higher as the total number of computations is higher than during the inference phase. %due to parallellization on the GPU. 
%The simulation speed on the GPU during both learning and inference phases is comparable, unlike the CPU platform, where learning takes longer time due to serialization of the computation.

\section{Real-time inference on user data}
%\section{System Implementation}
\label{preprocess}

We used the CUDA based SNN described in the previous section, %network using NVIDIA's CUDA parallel computing platform 
to design a user interface that can capture and identify  the images of digits written by  users in real-time from a touch-screen interface. The drawing application to capture the digit drawn by the user is built using OpenCV, an image processing library \cite{OpenCV}.
%The network is capable of running on any device with a CUDA-enabled GPU. 
The captured image from the touch screen is pre-processed using standard methods similar to that  used to generate the MNIST dataset images \cite{mnist}. We convert the user drawn images to the required format which is a grayscale image of size $28\times28$ pixels. %Fig.~\ref{fig:setup} shows the sample of 500 handwritten images previously mentioned. 
%The user inputs were collected by using a touch screen and 
The network is implemented on the NVIDIA GTX 860M GPU which has $640$ CUDA cores. The preprocessing phase  %(Fig \ref{fig:setup}) 
takes  about %$25\,$ms, 
$15\,$ms and this image is then passed to the trained  SNN for inference.
%The entire system is successfully able to identify the non-MNIST images of digits written by different users. %of the The average time for the neural network to predict an input image took about 425ms while using this mobile GPU.
The CUDA process takes about $300\,$ms to initialize the network in the GPU memory, after which the network simulation time depends on the presentation time $T$ and the time step interval $\Delta t$.

\begin{figure}[!htb]
	\centering
\includegraphics[width=1\linewidth]{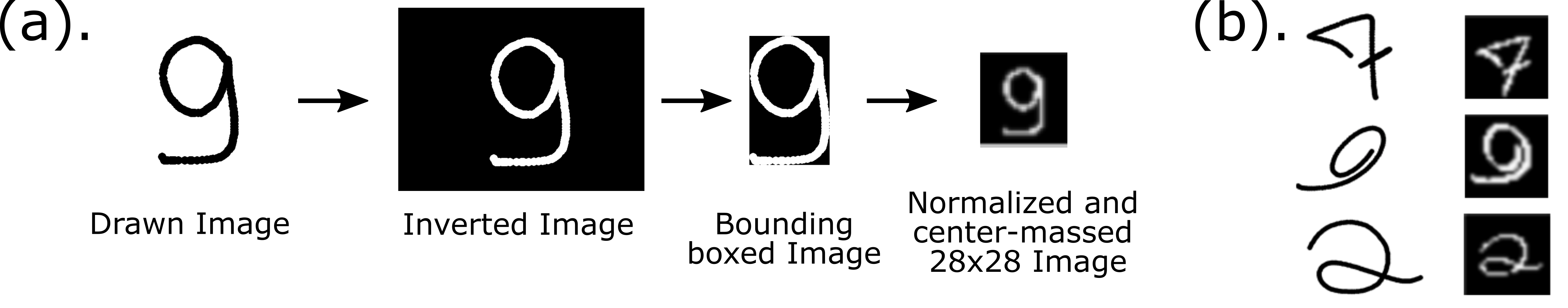}
	\caption{(a) Outline of the preprocessing steps used to convert the user input to a $28\times28$ image that is fed to the network, (b) Examples  of user input (left) and the pre-processed $28\times28$ pixel images fed to the SNN (right).%, (c) Collected sample of 500 handwritten digits from various users. 
	}    
	\label{fig:setup}
\end{figure}

\subsection{Image Preprocessing}
%Before the network can infer a digit written by the user, the entry must undergo simple preprocessing steps in order to create an input image that is similar to the ones found in the MNIST dataset.
Fig.~\ref{fig:setup}(a) shows the  preprocessing steps used to create the input signal to the SNN from  the captured image and Fig.~\ref{fig:setup}(b) shows some sample pre-processed images. %We first  apply an elastic distortion to vertically align the digit, and   then resize it  maintaining its aspect ratio.  The digit is placed by its center of mass, and  a border is applied to ensure that the final image is of size $28\times28$ pixels.
The image captured from the user is first binarized by thresholding and cropped to remove excess background. The image is resized to $20$ pixels along its longer dimension, while maintaining its aspect ratio. Thereafter, the resized image is placed in a $28\times28$ bounding box such that the image's center of mass coincides with the center of the bounding box. Finally, the image is passed through a blurring filter to create gray-scale images similar to the ones in the MNIST dataset.

%\begin{figure}[!htb]
%	\centering
%	\includegraphics[width=0.7\linewidth]{collectedDigits_small.png}
%	\caption{Collected sample of 500 handwritten digits from various users. 
%	}    
%	\label{fig:setup}
%\end{figure} 

\section{Results} 
\label{res}
We trained the network on the MNIST training data-set consisting of $60,000$ images, for $20$ epochs.  Our network achieves an error of $0.2\%$ on the  training set and   $1.94\%$ on the test set with a time step   of $\Delta t=0.1\,$ms when the network is simulated for $T=100\,$ms. %accuracy of $99.80\%$; 
%error of   $0.20\%$ and when  the  $10,000$ image test set is presented to the trained SNN,  we measure an %accuracy of $98.16\%$ error of $1.94\%$. This is with a time step interval of $\Delta t=0.1\,$ms when the network is simulated for $T=100\,$ms.
 Table~\ref{Table:acc} lists the state-of-the-art networks (ANN and SNN) %case second generation network and the spiking networks 
for the MNIST classification problem. It can be seen that though these networks have classification accuracies exceeding $99\%$,  they  use  more than $7\times$ the number of parameters compared to our network, which is designed to simplify the computational load in developing real-time system. %Thus, our network is designed to simplify the computational load in developing a real-time system.

\begin{table}[h]
\caption{Comparison of our SNN with state-of-the-art}\centering
\begin{tabular}{ |p{4.cm}| p{2.1cm}| p{1.2  cm}| }
\hline
 Network  and learning algorithm & Learning synapses &  Accuracy \\ 
 \hline
%	 ANN (LeNet-5) \cite{LeNet5} & $331,984$ & $99.05\%$ \\  
% GCNN (LeNet-5 + Gabor filters) \cite{gcnn} & $331,984$ & $99.32\%$ \\
%MCDNN (Multi-column Deep NN) with elastic distortion\cite{mcdnn} &$1,565,600$&$99.77\%$\\
%DNN with DropConnect, images cropped and rotation scaled
Deep Learning \cite{Dropconnect} & $2,508,470$ &$99.79\%$\\
 %SNN, with STDP \cite{diehl2015unsupervised} & $5,017,600$ & $95.0\%$\\
%SNN, with Backpropagation \cite{snn_bp} & $302,000$ & $98.77\%$ \\
 %Spiking ConvNet trained in non-spiking neurons
 ANN converted to SNN\cite{Diehl} & $1,422,848$ & $99.12\%$\\
 
%Convolution with back-propagation in spike domain 
4-layer convolution SNN
\cite{snn_bp} & $581,520$ & $99.31\%$\\
%4-layer convolution SNN
%\cite{snn_bp} & $568,960$ & $99.31\%$\\

%  Spiking ConvNet \cite{diehl2015fast} & $412,608$ & $99.12\%$\\
SNN, with NormAD (this work) & $81,120$ & $98.06\%$\\
%ANN, with Backpropagation (this work) & $81,120$ & $98.20\%$\\
 \hline
\end{tabular}
\label{Table:acc}
\end{table}

\begin{figure}[!htb]
\centering
%  \begin{minipage}[h]{0.24\textwidth}
% 	\centering
%     \includegraphics[width=1\linewidth]{accuracy_T.pdf}
%     %\includegraphics[width=1\linewidth]{test_accuracy_1ms.png}
%     %\includegraphics[width=1\linewidth]{acc_vs_T.png}
%      %\includegraphics[width=1\linewidth]{accuracy.png}
%   \end{minipage}
%    \begin{minipage}[h]{0.24\textwidth} \centering \includegraphics[width=1\linewidth]{real_time_sys.pdf}
%    %\includegraphics[width=1\linewidth]{PROCESSING_TIME_RESULTS.png}
%        \end{minipage}
%\includegraphics[width=0.65\linewidth]{test_accuracy_1ms.png}
\includegraphics[width=0.99\linewidth]{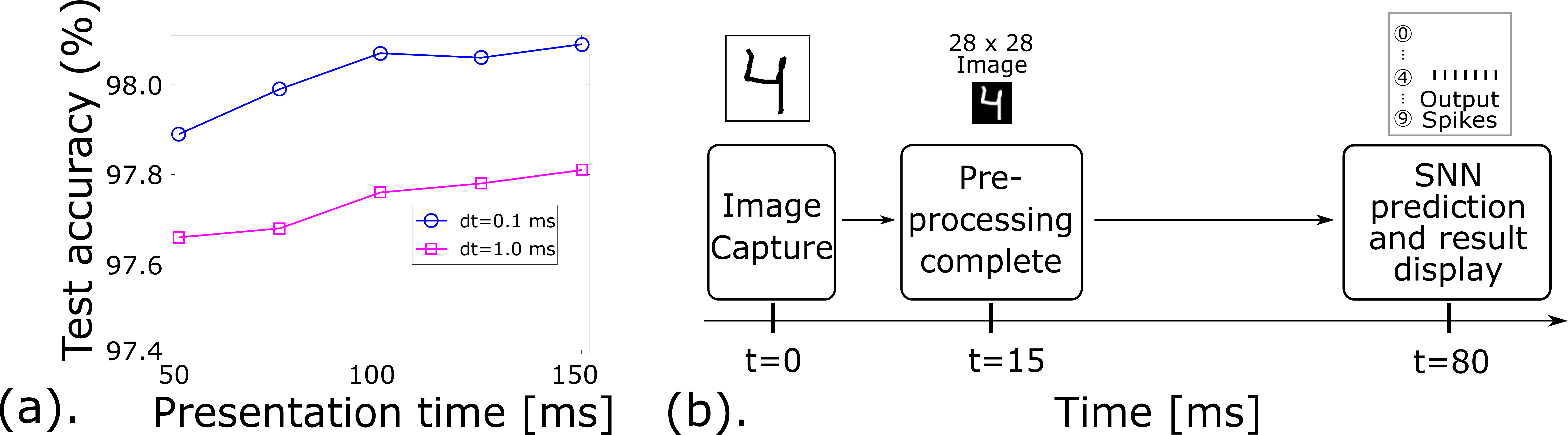}
       	\caption{%\textbf{(a)} Accuracy measurements  and \textbf{(b)} time needed to simulate the SNN as a function of the presentation time, with integration  time steps of $\Delta t=1\,$ms and $\Delta t=0.1\,$ms for 500 images captured through the touch-screen interface. The time needed to initialize the network in the GPU memory was approximately $400\,$ms.  
(a). MNIST test-set accuracy as a function of presentation time and the integration time step $\Delta t$. (b) Various stages of classifying a user's input: the image pre-processing takes $15\,$ms and the $75\,$ms SNN  emulation is completed in real-time. %\color{red}The accuracy trend for the images captured from the touch-screen interface is comparable to  the MNIST test data-set.\color{black}
}
       \label{fig:graphresults}
\end{figure}

If the integration time step interval used during inference  is $1\,$ms (i.e., approximating the neuronal integration) instead of $0.1\,$ms, the MNIST test  error increases only by about $0.4\%$ (see Fig.~\ref{fig:graphresults}(a)), but there is a $10\times$ reduction in the  processing time. Hence, for our touch screen based interface system we simulate the SNN with $\Delta t$ of $1\,$ms to infer the users' digits. When each digit is presented for  $T=75\,$ms, the network can be simulated in an average wall clock time of $65\,$ms, making real-time processing possible (Fig.~\ref{fig:graphresults}(b)).
%but the $T=100\,$ms SNN simulation can be completed within an average duration of  $77\,$ms (compared to $750\,$ms when $\Delta t=0.1\,$ms), making real-time processing possible.%i.e. a $10\times$ reduction in the simulation time.
%%%%%%
%%%%%%%%%%%%%%%%%
%Though state of the art deep convolution networks \cite{LeNet5,mcdnn}   have achieved %over $99\%$ accuracy, 
%less than $1\%$ error, they have in excess of $300,000$ trainable synaptic weights. In comparison, our network uses only  about $82,000$ trainable weights, i.e. nearly $4$ times fewer learning parameters.%(Table \ref{Table:acc}). %We also study the digit-wise accuracy of the network in correctly identifying the $10$ digits; it can be seen that the straight-forward  choice of the $8$ edge detection kernels in the pre-processing layer is sufficient to guarantee similar accuracy metrics (with a standard deviation of $2\%$) for the $10$ digits.
% Table \ref{Table:acc} lists accuracy and parameter size of our network with the different networks reported in the literature.
% \begin{figure}[!h]
% 	\centering
% 	\includegraphics[width=0.9\linewidth]{Error_no_wta.pdf}
%     	%\includegraphics[width=\linewidth]{accuracy_wo_dt_shift.pdf}
% 	\caption{Test and training set error percentage during $20$ training epochs. During each epoch,  all $60,000$ images from the MNIST training data-set is presented and the weights  updated after every image presentation. The test error is measured on the $10,000$ images of the test set. The error on the test set was $1.84\%$ and $0.20\%$ on the training set.}
% 	\label{fig:accuracy}
% \end{figure}
We tested the network's accuracy  with $\Delta t=1\,$ms  on a  set of $500$ handwritten digits collected from various users through our user-interface system. %Fig.~\ref{fig:graphresults}(b) shows the accuracy  as a function of the presentation time. 
At $T=75\,$ms, we measure an accuracy of $97.4\%$ on our set of $500$ captured images, while on the MNIST test-set it was $97.68\%$.
%The network was also tested with a sample set of $500$ handwritten digits collected from various users through our user-interface system. Among these $500$ images, we measure an error of $2.8\%$ when using a time step of $0.1$ms and a $3\%$ error when using a time step of $1$ms, for $T=100\,$ms (Fig.~\ref{fig:graphresults}). 
The slight loss in performance compared to the MNIST dataset is attributed to the deviations from the statistical characteristics of the captured images compared to the MNIST dataset.  % presents the measured  classification accuracy and the network processing time as a function of image presentation  time interval and the integration time step. 

\section{Conclusion}
\label{concl}
We  developed a simple three-layer spiking neural network that performs spike encoding, feature extraction, and classification.  
All information processing and learning within the network is performed entirely in the spike domain. With approximately $7$ times lesser number of synaptic weight parameters compared to the state of the art spiking networks, we show that our approach achieves classification accuracy exceeding $99\%$ on the training set of the MNIST database and  $98.06\%$ on its test set. 
The trained network implemented on the CUDA parallel computing platform is also able to successfully identify digits written by  users in real-time, demonstrating its true generalization capability.

We have also demonstrated a general framework for implementing spike based neural networks and supervised learning with non-local weight update rules on a GPU platform. At each time step, the neuronal spike transmission, synaptic current computation and weight update calculation for the network are all executed in parallel in this framework. 
Using this GPU implementation, we  demonstrated a touch-screen based platform for real-time classification of user-generated images.
%user-interfaced implementation of the system that can take inputs from  users and predict the label of the image in real-time. %This work will be further extended to incorporate online learning and inference with real users' data.
%, resulting in close to $6\times$ speed up for learning execution time compared to a single core CPU implementation. 

%\section*{Acknowledgment}
%\section*{Acknowledgment}
%This research  was supported in part by the CAMPUSENSE project grant from CISCO Systems Inc, the Semiconductor Research Corporation, McNair Scholars program and the National Science Foundation grant 1710009. 
\bibliographystyle{IEEEtran}
\bibliography{ref} 

\end{document}